\documentclass[letterpaper, 10 pt, conference]{ieeeconf}  

\IEEEoverridecommandlockouts                              

\usepackage{multirow}
\usepackage{graphicx}
\usepackage{svg}

\title{\LARGE \bf
Dynamic Risk Assessment Methodology with an LDM-based System for Parking Scenarios
}

\author{Paola Natalia Cañas$^{12}$, Mikel García$^{12}$, Nerea Aranjuelo$^{12}$, Marcos Nieto$^{1}$, Aitor Iglesias$^{1}$ and Igor Rodríguez$^{2}$ 
\thanks{Funded by the European Union from the Horizon Europe programme under grant agreement 101076868 (AWARE2ALL), and by the Basque Government  from the program ELKARTEK 2021 under project Autoev@l.}
\thanks{$^{1}$Fundación Vicomtech, Basque Research and Technology Alliance (BRTA),
        Donostia - San Sebastian (Spain)}%
\thanks{$^{2}$Universidad del País Vasco (UPV/EHU),
         Donostia - San Sebastian (Spain)}%
}

\begin{document}

\maketitle
\thispagestyle{empty}
\pagestyle{empty}

\begin{abstract}

This paper describes the methodology for building a dynamic risk assessment for ADAS (Advanced Driving Assistance Systems) algorithms in parking scenarios, fusing exterior and interior perception for a better understanding of the scene and a more comprehensive risk estimation. This includes the definition of a dynamic risk methodology that depends on the situation from inside and outside the vehicle, the creation of a multi-sensor dataset of risk assessment for ADAS benchmarking purposes, and a Local Dynamic Map (LDM) that fuses data from the exterior and interior of the car to build an LDM-based Dynamic Risk Assessment System (DRAS). 

\end{abstract}

\section{INTRODUCTION}

Modern vehicles are equipped with Advanced Driver Assistance Systems (ADAS) to warn the driver of potential hazards, provide decision-making support, and even replace the driver in simple tasks. However, parking is complex also for humans and has not yet been solved by automated functions, which can work only under certain circumstances (good painted lines or absence of curbs, cobblestones or uneven surfaces). Current parking assistance systems are based on proximity sensors or radars that act as a naive assistant for the driver. They perceive near objects without recognising them or assigning an individual risk. To be more competent, the systems should be able to establish when, what to warn and with what urgency. 

One key element in driver assistance systems is, actually, the driver. To advance toward warning the driver of actual unknown or not conscious risks, they must be included in the scene and considered by the ADAS. From mildly alerting about known risks (the driver acknowledges the obstacle approaching) to act in case of high unknown risks (the driver can not see or is not aware of them).

In addition to the driver, another essential element for risk assessment is the vulnerable road users involved in the risk (e.g. nearby pedestrians). The behavior of pedestrians has been studied in the context of potential collisions with vehicles, but typically only in scenarios where vehicles are moving forward \cite{Dafrallah}. Many variables differ in a parking scenario, including the driver’s visibility and vehicle speed. Our research studies new conditions, hazards, and behaviors of key elements in the scene and designs a risk assessment procedure accordingly.

The primary goal of this paper is to extend research on risk estimation in parking scenarios by taking into account additional variables: driver perception and relative positions of pedestrians surrounding a vehicle. This includes identifying whether the driver is looking at any rear-view mirrors and detecting potential obstacles using sensors such as LiDARs and cameras. The fusion of these two perceptions is achieved through an LDM (Local Dynamic Map), where the driver has not been considered until now. The main contributions of this paper are:

\begin{itemize}
    \item A methodology for a dynamic risk assessment in parking scenarios, defining several parameters and making measurements.
    \item A dynamic risk scale, depending on the interior and exterior situation of the vehicle.
    \item A dataset compliant with the risk methodology and scale presented. 
    \item An approach of an LDM-based dynamic risk assessment system to fuse the interior and exterior situation of the vehicle to showcase the proposed methodology of risk assessment.
\end{itemize}

\section{STATE-OF-THE-ART}

Although autonomous driving technology has made significant progress, it still faces many challenges. One of the most pressing issues is the risk assessment of autonomous vehicles, which has garnered significant attention from the public, industry and regulation bodies. Some research has been conducted on risk estimation in vehicle-pedestrian interactions in regular traffic situations like a crosswalk. For example, Nie et al. \cite{Nie} focused on determining the safety boundary based on distance (known as the “safety envelope”) for conflicts between vehicles and pedestrians. Their study used physiological signal measurements and kinematic reconstruction to understand how pedestrians naturally avoid harm. Their safety envelopes were generated by taking into account the interactions between the pedestrian and the vehicle.
In another study, Kolekar et al. \cite{Kolekar} define a Driver’s Risk Field (DRF) model that estimates a driver’s perceived risk by representing their belief about the probability and consequences of an event. By multiplying the DRF by the consequence of the event, the model provides an estimate of the driver’s perceived risk. It means they define a methodology to dynamically define risk zones and scale for collisions to any object based on the driver's perception. In other studies, the collision risk depends on the vehicle's stopping time \cite{SGI}. Dafrallah et al. \cite{Dafrallah} defined four possible risk zones: the danger zone, the alert zone, the warning zone and the safe zone.

The definition of risk zones for parking use cases can be briefly seen in the ISO 22840:2010 standard \cite{ISO22} for Extended-Range Backing Aids (ERBA), which establishes minimum functionality requirements for detecting and informing the driver about relevant obstacles; and in ISO 17386:2004 \cite{ISO17} standard for Manoeuvring Aids for Low-Speed Operation (MALSO), which specifies minimum functionality requirements for detecting and informing the driver about relevant obstacles within a short detection range. These standards define monitoring zones classified as rear 1, rear 2, with the latter being slightly wider than the former and the rear corner covering the rear corners of the vehicle. Another thing to note about these standards is that they require parking assistance systems to detect a range of 1m to 5m from the car rear, distances that are taken into account in this study.


A simpler approach to risk assessment is currently integrated into many commercial vehicles to warn drivers about obstacles during parking manoeuvres. They commonly use ultrasonic sensors \cite{bosch} to measure the distance from the bumper to any nearby obstacles. A closer distance is considered more urgent and requires immediate attention, while a longer distance is less urgent. The situation's urgency is typically communicated to the driver through an audible warning sound.

Perception data may be consumed by different components of a vehicle in order to understand its surroundings and act accordingly. A Local Dynamic Map (LDM) constitutes a key component in which information from multiple data sources can be stored and consumed in real-time. In the SAFESPOT project specification document \cite{SAFESPOT} they defined the main four-layer model that defines a LDM, in which the stored data is structured and assigned to a layer depending on its dynamics and time span. The original four layers consist of a static layer (map/road data), a quasistatic layer (traffic signs, landmarks, etc.), a semi-dynamic layer (traffic light phase, traffic congestion, etc.) and a dynamic layer (vehicles, pedestrians, etc.). Different works have developed and further evolved this initial model, in \cite{R-LDM} they extended the LDM concept by using a relational database instead of the SQL database used in the original SAFESPOT implementation. In \cite{ILDM} a graph database is used in combination with the data schema of the OpenLABEL standard, focusing in an interoperable implementation of a LDM. Other works such as \cite{Six-Layer} extend the original four-layered model to a six-layered model making a clearer distinction of layers for dynamic information and including a digital information layer for Vehicle-to-Everything (V2X) communications. In our proposal, an additional layer is considered in order to store relevant information about the interior of a vehicle, which can be provided by a Driver Monitoring System (DMS) or other inside monitoring solutions.

\section{DYNAMIC RISK ASSESSMENT METHODOLOGY}

\subsection{Scenario Description}

\begin{figure}[htbp]
  \centering
  \includegraphics[width=2in]{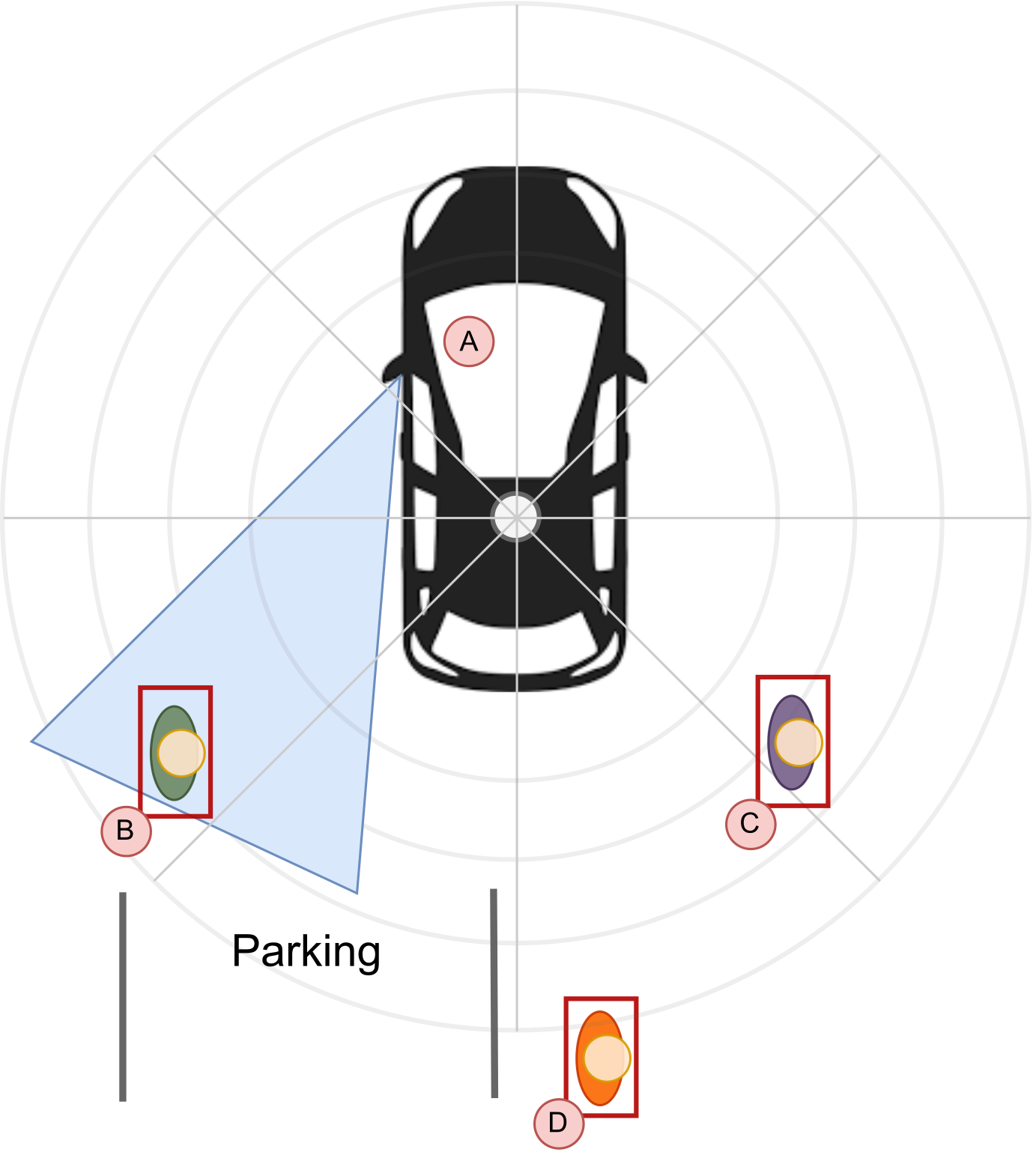}
  \caption{Diagram of different scenarios in a parking maneuver}
  \label{fig:scenarioDiagram}
\end{figure}
As mentioned earlier, the methodology presented in this paper considers parking scenarios. For the purposes of this discussion, Subject A will refer to the driver and other Subjects will refer to pedestrians. In this scenario, Subject A wants to reverse into a parking spot and relies on the three rear-view mirrors to visualize the area behind the car and identify potential risks of hitting or running over a person. Subject B is a pedestrian who is unaware of Subject A’s intended maneuver and is walking on a trajectory that brings them close to the car, with the possibility of being hit.

Different risks can be present in this scenario, partly dependent on the distance between Subject B and Subject A’s vehicle. We propose a dynamic risk scale that also depends on Subject A’s awareness of the situation. In the diagram presented in Fig. \ref{fig:scenarioDiagram}, Subject A is looking at the left rear-view mirror while Subjects B, C, and D are walking around. Intuitively, Subjects B and C are at the same distance from the car and represent more risk than Subject D, who is farther away. However, when we incorporate the driver’s awareness variable, this changes. Since Subject A can see the left side of the car, it is assumed that Subject's B presence is acknowledged. Therefore, presence of Subject C should represent more risk that Subject B's.

\subsection{Risk zones}\label{risk_zone}
\begin{figure}[htbp]
  \centering
  \includegraphics[width=2.5in]{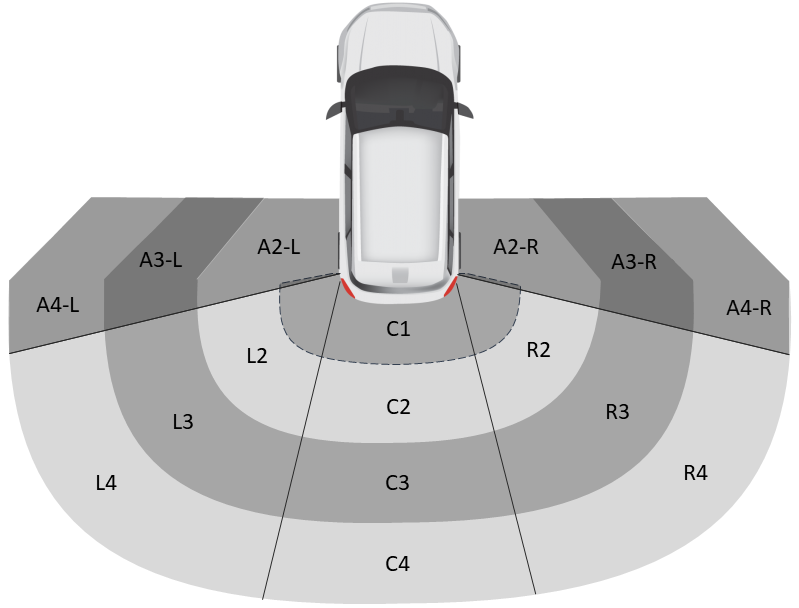}
  \caption{Diagram of risk zones}
  \label{fig:riskzones}
\end{figure}
The diagram of Fig. \ref{fig:riskzones} illustrates the zone defined and taken into consideration in this study. The zone extends 4m from the car rear-bumper countour and is divided into smaller sections. Each section is labelled with a number indicating its distance from the vehicle and a letter representing the area visible to the driver when looking in the corresponding rear-view mirror: "L" (left), "R" (Right), "C" (Center). As the car is assumed to be moving in reverse, the risk zone covers primarily the rear area of the car. Additionally, there are sub-zones denoted with "A" that expand the risk zone to consider the two lateral areas of the back of the vehicle; these can be optionally considered if a broader risk estimation wants to be done. The “A” sub-zones have a unique shape with cuts at the edges to account for pedestrians who, while following a normal trajectory towards the car, either enter directly into the driver’s field of vision or are out of danger by being located in front of the car (assuming that it is moving backwards) rather than at its rear. It is important to note that the risk zone depicted in the diagram is relative to the car and "moves" with it. As a result, a stationary pedestrian may pass through different risk zones as the car and its associated risk zones move in reverse.

\subsection{Risk Levels \& Variables}
\subsubsection{Risk Levels Definitions} \label{risk_levels}
\begin{figure}[htbp]
  \centering
  \includegraphics[width=1.5in]{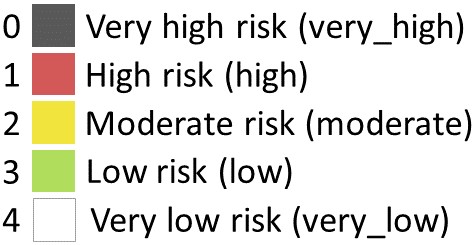}
  \caption{Risk levels scale}
  \label{fig:levelColors}
\end{figure}

Risks levels are determined by two factors: the distance between the vehicle and the pedestrian (determined by the sub-zone itself) and the driver’s estimated awareness of the pedestrian’s presence. The levels defined are associated with colors and shown in Fig.~\ref{fig:levelColors}.
The distance-associated risk level is calculated using Time To Collision (TTC) metric. TTC estimates the time until a collision may occurs based on the current velocity (speed and direction) of both subjects: Subject A and B. In this research, for simplicity, Subject A’s velocity is fixed at 5km/h moving backwards, while Subject B’s velocity is not considered. Instead, our focus is on Subject B’s position within the risk zone at a specific time, independently of their previous or future position. This information could be important for researchers working on algorithms for pedestrian detection.

With time and speed defined, the TTC metric is calculated. Now, it is possible to assess the risk of collision based on this metric. Intuitively, the closer the subjects are to each other, the higher the risk. However, the exact scale of this closeness is what we aim to define. The highest risk scenario would be one in which immediate action is required to prevent a collision, but the driver is unable to physically react in time due to recognition and reaction time delays. The total driver reaction time is typically defined as 1.5 seconds, during which the vehicle can travel approximately two meters before coming to a complete stop. Therefore, areas within two meters (all sub-zones labelled with "2") of the vehicle are considered to have a high risk if the driver is aware of this zone. An additional distance of one meter from the high-risk zone is defined as having a moderate risk. This is because, while it is farther from the vehicle, it is still relatively close to the high-risk zone. Sub-zones that are four meters away from the vehicle are considered to have a low risk. Areas outside of the defined risk zone (farther than four meters and in front of the car) are not considered as they have a very low risk. Finally, we have one more level of risk to define: very high risk. This level is assigned to cases where the driver is unable to perceive the danger and therefore cannot react in time. 

\begin{figure}[htbp]
  \centering
  \includegraphics[width=3.3in]{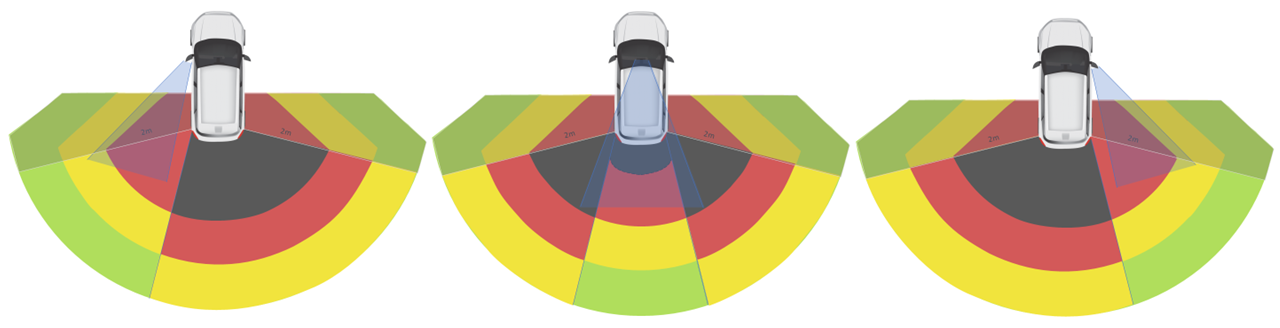}
  \caption{Risk assessment dependant on distance and driver awareness}
  \label{fig:riskTodas}
\end{figure}

In this research, it is proposed that if Subject A is unaware of the danger, the risk is even higher. To calculate the areas of non-visibility for an average driver in an average personal vehicle, the limits of the driver's vision when looking at each of the rear-view mirrors were estimated by asking someone in the driver's seat if a pedestrian was visible or not in a specific point. This determined the zones represented in Fig. \ref{fig:riskzones} and its division into three (left, centre, and right), as explained above. For the next variable: driver awareness, it was assumed that when Subject A looks at a rear-view mirror, they consciously examine the space to make a manoeuvre and can detect danger if it appears. As such, the corresponding area under Subject A's control or understanding will maintain its risk levels. The other two zones will increase their risk levels: going from high to very high risk or from moderate to high risk. This dynamic change in risks can be seen in Fig. \ref{fig:riskTodas}. 
Additionally and related to the driver's visibility, the one meter closest to the rear of the car sub-zone (C1) is categorized as having a very high risk due to limitations in visibility from the centre rear-view mirror because of the morphology of the car, causing that pedestrians, especially infants, may not be visible in this area.  

\subsubsection{Variables and Parameters}
To define the risk zones and their associated levels, we took into account several variables and parameters. These include:
\begin{itemize}
\item Vehicle velocity during parking manoeuvres: The Field Measurement of Naturalistic Backing Behaviour \cite{ntsha} study of the National Highway Traffic Safety Administration (NHTSA) collected data in real-world driving conditions. It was found that, except for extended backing manoeuvres, backing speeds averaged around 4.8 km/h. Additionally, the maximum backing speed for young drivers was faster than that of elderly drivers and males backed faster than females. To account for extreme cases and individuals with longer reaction times, we rounded this value up to 5 km/h.

\item Rear-view mirror visibility and vehicle blind spots: These are essential factors to consider when assessing risk in parking scenarios. The visibility offered by the rear-view mirror can vary depending on the mirror itself and the driver’s preferred position or angle. Some commercial vehicles include modifications on the mirrors to eliminate blind spots; also, some drivers rather see the rear of the car in the mirror and some others follow the recommendations of driving agencies and set the mirror to have a wider view of their surroundings. To address this issue in our study, we measured the visibility and blind spots of our test vehicle, a third-generation Toyota Prius, to define the right and left boundaries of the risk zones: the boundaries of zones marked with “L”, “R”, and “C”. While there may be some variation among different commercial cars regarding visibility and blind spots, we believe these differences won’t significantly affect the risk zones and sub-zones. Implementing this approach in larger vehicles like trucks would require additional measurement and definition phases.

\item Driver's reaction time: Research on driver reaction time has shown that it varies depending on factors such as the situation, the driver's age, and the anticipation of an incident. For example, Olson \cite{olson} found that for a "straightforward" situation, 90-95\% of reaction times were between 0.75s and 1.5s." In this study, 1.5s will be considered the average reaction time used to define the risk scale. This reaction time includes the braking time, which at higher speeds should be considered. However, being a parking use case and at a low speed, it is discarded.

\end{itemize}

\subsection{Dataset}
Once the methodology for risk assessment is defined, we added a practical aspect to it by creating a dataset that includes scenes associated with the risks introduced in this research. This serves as validation data for a risk assessment system (DRAS) that is also developed and will be explained in further sections. In other words, this data is compliant with the risk methodology described and will validate that the way of defining and estimating risk of this research is feasible using state-of-the-art algorithms and today's technology. This data (to be published), can serve for benchmarking purposes for other research on risk assessment algorithms, and to test the system performance in risk estimation instead of individual algorithms for detection, such as pedestrian detectors.

This dataset includes data from various sensors and risk classifications (classes) with an additional label about the mirror the driver is looking at, allowing parking-aid systems to be tested in terms of risk assessment in parking scenarios using our defined risk zones and scale. 

To capture the recordings, we equipped the car with several sensors to record comprehensive data. A rear camera was installed to record the view behind the car, while a face camera in the cockpit monitored the driver’s gaze. A Velodyne HDL-32E LiDAR on top of the car also provided point clouds of the scene for improved pedestrian detection and distance estimation. This dataset offers multi-sensor data, which is important for the validation of the detections through different algorithms, in this case, pedestrian detection with LiDAR or camera.

\begin{figure}[htbp]
  \centering
  \includegraphics[width=3.3in]{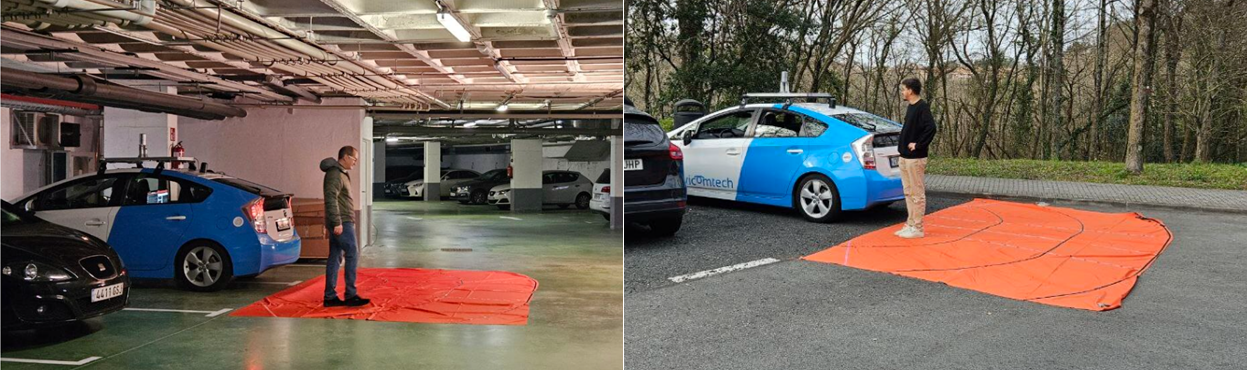}
  \caption{Images of the recording process of the dataset (indoor and outdoor scenarios)}
  \label{fig:rec}
\end{figure}

To ensure accurate representation of risks during recording, we marked risk zones on a large piece of fabric and placed it on the ground. This helped participants know where to position themselves, be consistent and mitigate errors in distances if the recording scenario changed. Some images showing how the recording process was done are shown in Fig. \ref{fig:rec}.  

To ensure the accuracy of the labels, the recordings are organized in a way that each video was saved under a folder corresponding to its specific risk level. This allowed for efficient and accurate labelling of the data. Since the detection of pedestrians and the assessment of risk are analyzed as independent events in time rather than as temporal events that depend on the previous state of the pedestrian, the recordings were performed with the car parked. People position in one risk zone and the driver looks at one mirror per video.

\section{LDM-BASED DYNAMIC RISK ASSESSMENT SYSTEM (DRAS)}
To demonstrate the dynamic risk definition in practice, a risk estimator system was developed that takes into account both the inside and outside of the vehicle. This system is based on an LDM and implements two algorithms: one for person detection and another for gaze detection. The dataset described in the previous section was used to validate this system. It is important to note that while the detection system (both detection algorithms) is necessary for testing the methodology of dynamic risk assessment in parking scenarios and for the construction of the LDM, it is not the main contribution of this work. The algorithms used can be interchangeable.

\subsection{Detection Algorithms}

\subsubsection{LiDAR-based Pedestrian Detection}\label{SCM}
Regarding the work on pedestrian detection in point clouds, we used a voxel-based CenterPoint \cite{CenterPoint} deep learning model. This model is able to detect and classify different types of 3D objects such as pedestrians, cars, and more. We used the implementation of MMDetection3D \cite{MMDet3D}. We trained CenterPoint with the nuScenes dataset \cite{nuScenes} without accumulating scans. NuScenes is a large-scale dataset for autonomous driving research containing multimodal sensor data collected from real-world traffic scenarios. Although the model can detect up to ten classes, we used detections of pedestrians and cars exclusively.

\subsubsection{Rear-view Mirror Gaze Detection}
Another algorithm that was implemented was a rear-view mirror gaze detector. It detects which of the three rear-view mirrors the driver is looking at. To develop this algorithm, the Driver Monitoring Dataset (DMD) \cite{dmd} was used. It contains video footage of people looking at nine predefined parts of a car, focusing on the classes for the left mirror, center mirror, and right mirror for the present work. Using transfer learning from ImageNet, a model was trained to perform the task of detecting the driver’s gaze accurately. The input for this model is an image and the output is the mirror the driver is looking at. 

\subsection{Local Dynamic Map}
The LDM has commonly been used for storing information about an ego vehicle or its near environment. As far as the authors of this work are aware, no LDM model or implementation has ever considered fusing the interior information of the vehicle by monitoring the driver or passengers. For this exercise that combines exterior and interior variables for a risk assessment, it was also decided to build an LDM and become the bridge between these two worlds, in which there will be information from the exterior (3D object detections) and the interior (driver's gaze). 

This risk assessment system relies on obtaining information from two different detection systems (driver gaze and object detection). A simple LDM implementation has been developed in this work using InfluxDB\footnote{https://www.influxdata.com/} as the database engine. InfluxDB is a time-series database (TSDB) optimized for real-time usage and querying. The LDM consists of two data sources that are constantly updated by parsing the data obtained from the detection algorithms into the LDM database. 

\subsection{Dynamic risk assessment component} \label{DRAC}
\begin{figure}[htbp]
  \centering
  \includegraphics[width=3in]{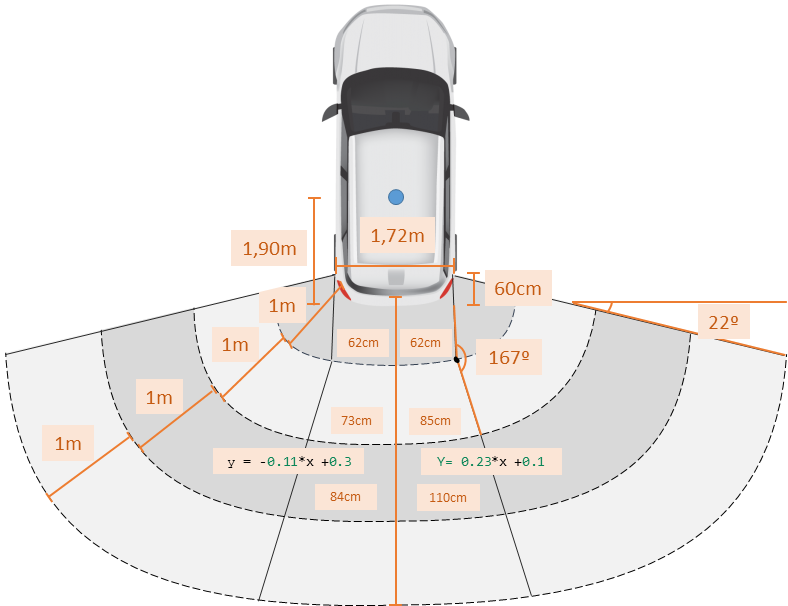}
  \caption{Measures of the risks zones to represent them into the virtual 3D world.}
  \label{fig:medidas}
\end{figure}

With both type of detections now possible: the position of the pedestrian in the virtual 3D world and the rear-mirror to which the driver is looking, a dynamic risk assessment component is developed to provide a risk value, with the risk scale presented in Section \ref{risk_levels}, based on these two variables. 

To identify in which risk zone is the pedestrian detected, they are represented into the virtual 3D world, the one in which the pedestrian detections through LiDAR are made. To determine whether a detected person is in an “l”, “c”, or “r” zone, line equations were estimated using two points from the division lines on the fabric: one equation for the left division line and one for the right division line as can be seen in Fig. \ref{fig:medidas}. If a given (x,y) point is above the left division line equation, it is considered to be in the “l” zone. If it is below this line and above the right division line, it is considered to be in the “c” zone. Otherwise, it is in the “r” zone.

The distances that define the risk zones are calculated with respect to the car rear-bumper contour, while the coordinates of the detected pedestrians are relative to the LiDAR (origin of the virtual 3D world coordinate system). A transformation is applied to pedestrians detection points (x,y) to change the origin from the center of the rear bumper projected to the ground (following the ISO-8855) to the position of the LiDAR. Additionally, only detections with x's with positive values and less then 6m (after the transformation) are processed. Outside that range it is considered to be out of the risks zones, therefore is a very low risk.

Once the risk zone of a pedestrian is assigned and the rear-mirror being monitored by the driver is detected, a risk is assessed following the methodology represented in Fig. \ref{fig:riskTodas}.

\section{RESULTS}

\subsection{Validation Dataset}
A dataset of 24 scenes recorded with a rear and driver camera (RGB images) and LiDAR (point cloud data) was achieved with the following distribution of material by classes (risks), by driver gaze and by scenario shown in Table~\ref{tab:distrib}. 
\begin{table}[]
\caption{\label{tab:distrib} Distribution of dataset material}
\begin{tabular}{l|ccc|ccc|c}
Scenario & \multicolumn{3}{c|}{Interior} & \multicolumn{3}{c|}{Exterior} & \multirow{2}{*}{Total (\%)} \\ \cline{1-7}
Gaze & left & center & right & left & center & right &  \\ \hline
risk 0 & 29 & 83 & 32 & 84 & 249 & 147 & 29\% \\
risk 1 & 63 & 33 & 60 & 203 & 175 & 265 & 37\% \\
risk 2 & 26 & 31 & 111 & 164 & 81 & 162 & 26\% \\
risk 3 & 38 & 0 & 0 & 34 & 46 & 55 & 8\% \\ \hline
Total (\%) & \multicolumn{3}{c|}{23\%} & \multicolumn{3}{c|}{77\%} & \multicolumn{1}{l}{} \\
\end{tabular}
\end{table}

\subsection{DRAS Performance}
\begin{figure}[htbp]
  \centering
  \includegraphics[width=3.3in]{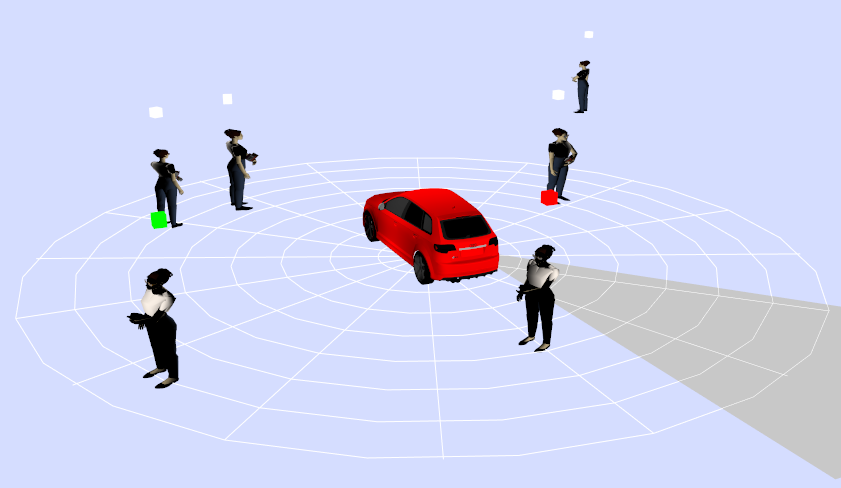}
  \caption{Visualization of the LDM-based Risk Assessment System (DRAS)}
  \label{fig:vis}
\end{figure}

The visualization of the DRAS being executed is shown in Fig. \ref{fig:vis}. In this image, the driver’s gaze is represented by a large grey triangle indicating the view when looking at the right mirror. Each detected person is depicted using a 3D model, with a colored cube above their head indicating their risk level. As can be seen, the person behind the car is almost 3m away, as determined by counting the 1m radio circles on the floor polar grid. Given their distance and position, it can be assumed that this person is in zone “c3”. With the driver looking at the right mirror, a high risk is assessed for this person, as shown in the right diagram of Fig. \ref{fig:riskTodas}. 
The system and its individual algorithms were tested using the created dataset. The results are shown in Table \ref{tab:acc}. The pedestrian detector achieved an accuracy of 92\% in calculating the risk zone of the person; this could be affected by measurement differences in the fabric and its representation in the virtual 3D world. The gaze estimator achieved a lower accuracy of 73\% on this dataset. When these predictions were combined in the dynamic risk assessment, the overall system accuracy was 83\%. This is a reasonably good metric for the task proposed in this paper: risk assessment for parking scenarios. 

As mentioned earlier, the algorithms used in the system are interchangeable and their individual accuracies are provided to estimate their influence on the system’s performance. For further implementation of a DRAS, a more accurate gaze estimator algorithm could be used.

\begin{table}[]
\caption{\label{tab:acc} Accuracy of the system and individual detectors.}
\begin{tabular}{l|ccc|ccc|c}
Scenario & \multicolumn{3}{c|}{Interior} & \multicolumn{3}{c|}{Exterior} & \multirow{2}{*}{Acc} \\ \cline{1-7}
Driver gaze & left & center & right & left & center & right &  \\ \hline
Ped. detector & 0.77 & 0.98 & 0.99 & 0.88 & 0.95 & 0.95 & \textbf{0.92} \\
Gaze detector & 0.55 & 0.34 & 1.0 & 1.0 & 1.0 & 0.45 & \textbf{0.73} \\
DRAS & 0.87 & 0.74 & 0.99 & 0.88 & 0.99 & 0.51 & \textbf{0.83}
\end{tabular}

\end{table}

\section{CONCLUSIONS \& FUTURE WORK}

This paper presents a methodology for dynamic risk assessment in parking scenarios that considers both exterior and interior factors. The approach involved taking measurements, establishing parameters to define risks zones and creating a risk scale. A DRAS was built with an LDM that serves as a hub for fused information of exterior and interior perception. It can also provide a real-time visualization of the scene. To test this methodology and the system, a dataset compliant with the described definition of risk was created. With the given results of this system that assesses risk in a parking scenario with an accuracy of 83\%, the feasibility of using the risk assessment methodology presented here with today's technology is proven. 

For the future, changing or improving the detection algorithms will enhance the system’s performance. New or personalized applications of this methodology can be produced by adjusting the risks zones: making measurements of the field of view of the mirrors or blind spots and distances from the vehicle rear-bumper. One adaptation that may be particularly relevant is the definition of this dynamic risk assessment for larger vehicles such as trucks. The possibility of extending the risk estimation to all the surroundings of the car, and not just the rear for parking scenarios, is also be considered.

\addtolength{\textheight}{0cm}   





\end{document}